\pdfoutput=1
\documentclass[11pt]{article}

\usepackage[preprint]{acl}

\usepackage{times}
\usepackage{latexsym}
 \usepackage{amsmath} 
 \usepackage{hyperref}

\usepackage[T1]{fontenc}
\usepackage[utf8]{inputenc}

\usepackage{microtype}

\usepackage{inconsolata}

\usepackage{graphicx}
\usepackage{booktabs}
\usepackage{multirow}
\usepackage{array}
\usepackage{amssymb}    % For \mathbb and other mathematical symbols
\usepackage{amsfonts}   % For additional mathematical fonts
\title{The Knowledge Microscope: \\ Features as Better Analytical Lenses than Neurons}

\author{Yuheng Chen,
    Pengfei Cao,
    Kang Liu,
    Jun Zhao\\
The Laboratory of Cognition and Decision Intelligence for Complex Systems, \\
Institute of Automation, Chinese Academy of Sciences, Beijing, China\\
School of Artificial Intelligence, University of Chinese Academy of Sciences, Beijing, China\\
\texttt{chenyuheng2022@ia.ac.cn, \{pengfei.cao,kliu,jzhao\}@nlpr.ia.ac.cn}}

\begin{document}
\maketitle

\begin{abstract}
Previous studies primarily utilize MLP neurons as units of analysis for understanding the mechanisms of factual knowledge in Language Models (LMs); however, neurons suffer from polysemanticity, leading to limited knowledge expression and poor interpretability. In this paper, we first conduct preliminary experiments to validate that Sparse Autoencoders (SAE) can effectively decompose neurons into features, which serve as alternative analytical units. With this established, our core findings reveal three key advantages of features over neurons: (1) Features exhibit stronger influence on knowledge expression and superior interpretability. (2) Features demonstrate enhanced monosemanticity, showing distinct activation patterns between related and unrelated facts. (3) Features achieve better privacy protection than neurons,  demonstrated through our proposed FeatureEdit method, which significantly outperforms existing neuron-based approaches in erasing privacy-sensitive information from LMs.\footnote{Code and dataset will be available.}.

\end{abstract}

\section{Introduction}
\label{section:Introduction}
Language Models (LMs) have demonstrated remarkable capabilities in storing and expressing factual knowledge \citep{TheC3, openai2024gpt4technicalreport,gemmateam2024gemma2improvingopen}. However, the underlying mechanisms remains unclear. Mechanistic interpretability of factual knowledge in neural networks aims to decompose these systems into interpretable units to understand how facts are stored and retrieved \cite{chen2024knowledgelocalizationmissionaccomplished}. The critical first step in this investigation is to identify the appropriate analytical units.
One mainstream approach is the neuron-based research method \citep{geva2021key-value,dai2022knowledge,chen2024journey}, which posits that LMs recall facts through multilayer perceptron (MLP) weights and conceptualizes the responsible knowledge storage units as \textit{knowledge neurons}.
Although using neurons as the research unit is highly intuitive, this kind of approach still has some notable limitations \citep{hase2023does,niu2024what,chen2024knowledgelocalizationmissionaccomplished}.

\begin{figure}
    \centering
    \includegraphics[width=\linewidth]{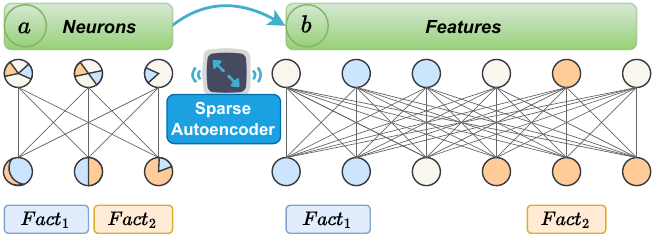}
\caption{Comparison of research units for factual knowledge mechanisms in LMs: (a) neurons and (b) features. Colors in neurons (or features) correspond to the facts they store, illustrating how specific facts are encoded in particular units.}
    \label{fig:introduction_figure1}
\end{figure}
In particular, a significant issue with neuron-based approaches is the phenomenon of polysemanticity \citep{bricken2023monosemanticity,Cunningham2023SparseAF}, where neurons  respond to mixtures of seemingly unrelated  facts. Intuitively, the number of factual knowledge items stored in LMs often exceeds the number of neurons\footnote{For example, Gemma-2 2B has $26\times9216 \approx 230k$ neurons but is trained on approximately 2 trillion tokens of data.}, necessitating that a single neuron must be associated with multiple facts.
As shown in Figure \ref{fig:introduction_figure1}(a), a fact may be dispersedly stored in a fragmented manner across numerous neurons, resulting in inseparable sets of neurons corresponding to $Fact_1$ and $Fact_2$. This fundamental characteristic leads to two challenges.

(1) \textbf{Limited Knowledge Expression:} 
Since factual knowledge can be dispersedly stored across numerous neurons, with some neurons potentially contributing only minimal information, these weak and distributed signals result in identified neurons having limited impact on knowledge expression.
(2) \textbf{Poor Interpretability:} The coupling of neurons representing different facts, makes it difficult to accurately describe the function of individual neurons, hindering our ability to gain deep insights into the mechanisms of factual knowledge.

\citet{bricken2023monosemanticity} suggests that polysemanticity arises from superposition, where neural networks represent more independent features through linear combinations of neurons. 
Their work demonstrates that Sparse Autoencoders (SAE) can effectively decompose neurons into interpretable features. 
As shown in Figure \ref{fig:introduction_figure1}, SAE transforms a ``low-dimensional'' neuronal space (a) into a ``high-dimensional'' feature space (b), making previously inseparable problems separable. This motivates us to explore whether such transformation could benefit the understanding of factual knowledge, where the transformed feature-level units might exhibit both stronger impact on knowledge expression and superior interpretability compared to their neuron-level counterparts.

Building on this foundation, we first investigate a preliminary question: Can neurons be effectively decomposed into features in the domain of factual knowledge, and is SAE a suitable technique for this decomposition? Then, we further explore three core research questions:

\textbf{Q1}: Can feature-based research methods address the dual challenges of Limited Knowledge Expression and Poor Interpretability?
\textbf{Q2}: Given that the key limitation of neurons lies in their polysemanticity, do features in the factual domain exhibit better monosemanticity?
\textbf{Q3}: Do features outperform neurons in downstream tasks?

Our investigation into these questions yields one preliminary finding and three core findings:

\textbf{Preliminary Finding} (\textbf{\textsection\ref{section:preliminary}}):  \textbf{SAE demonstrates superior effectiveness in decomposing neurons into features compared to other methods}, making them suitable alternative research units for studying factual knowledge mechanisms.

\textbf{Core Findings:}
(1) \textbf{Features as research units address the challenges of Limited Knowledge Expression and Poor Interpretability (\textsection\ref{subsection:Post-MLP Features Have the Strongest Impact on Knowledge Expression} and \textsection\ref{subsection:Features Demonstrate Superior Interpretability Compared to Neurons})}.  Through comparison of different modules (post-attention residuals, MLP activations, and post-MLP residuals), we find that features consistently show better interpretability than neurons, with post-MLP residual features having the greatest impact on knowledge expression.

(2) \textbf{Features exhibit stronger monosemanticity than neurons (\textsection\ref{subsection: Features Exhibit Stronger Monosemanticity}}).
Features strongly activate only when encountering related facts and remain inactive for unrelated ones, resulting in distinct separation in their activation distributions. In contrast, neurons lack such separation, indicating their susceptibility to activation by unrelated facts and weaker monosemanticity, while features better align with the ideal scenario in Figure \ref{fig:introduction_figure1}(b).

These two findings complement each other: the superior interpretability of features  naturally arises from their stronger monosemanticity property.

(3) \textbf{Feature-based method demonstrates superior performance in knowledge erasure for privacy protection (\textbf{\textsection\ref{section:Privacy Protection}})}. 
We propose \textbf{FeatureEdit}, the first feature-based model editing method, and evaluate it on our newly constructed privacy knowledge dataset \textbf{PrivacyParaRel}. Compared to neuron-based approaches, FeatureEdit achieves higher success rates in erasing privacy-sensitive knowledge from LMs while maintaining better generalization across semantically equivalent rephrase queries. Moreover, it causes less damage to the model's general capabilities.

\section{Dataset, Models and Evaluation metrics}
\label{section:Dataset, Models and Evaluation metrics}
Our experiments leverage Gemma Scope \citep{lieberum-etal-2024-gemma}, a comprehensive suite of SAEs trained on Gemma-2 models \citep{gemmateam2024gemma2improvingopen}. We study the 2B and 9B variants as they have SAEs for all layers.  Regarding the dataset, consistent with other neuron-based methods \citep{dai2022knowledge, chen2024journey}, we employ the ParaRel dataset \citep{elazar2021measuring-dataset}. For details to the dataset, see Table \ref{appendix:tab:relation_examples} in Appendix \ref{section:appendix-dataset}.

We introduce two evaluation metrics most frequently used in this paper. (1) $\Delta Prob$, the decreasing value of answer probability after features/neurons ablation, which assesses the impact of knowledge storage units on knowledge expression. For neurons, we directly set their activations to zero. For features, we aim to perform a similar operation. However, since features do not exist explicitly in LMs, we propose a reconstruction-based method (detailed in Appendix \ref{subsection:appendix:Feature Ablation Proces}). Briefly, given an activation ($\mathbf{h}$), we obtain its corresponding features, set target features to zero, reconstruct the activation ($\mathbf{h}'$), and replace $\mathbf{h}$ with $\mathbf{h}'$. We then measure the change in the probability of the correct answer before ($Prob_b$) and after ($Prob_a$) ablation: $\Delta Prob = \frac{Prob_b - Prob_a}{Prob_b}$.

(2) $IS$, Interpretability Score, which measures the interpretability of features. We  modify the method from \citet{bills2023language} to adapt it to our task (detailed in Appendix \ref{subsection:appendix:Autointerpretation Protocol}). Briefly, for features or neurons, we ask Large LMs (this paper uses gpt-4o-mini) to predict their activations. The correlation between the model's predicted activations and the actual activations is the interpretability score.

\section{Preliminary Experiment}
\label{section:preliminary}
\subsection{SAE Shows Superior Performance}
We first address the preliminary question raised in \textsection\ref{section:Introduction}: whether neurons can be decomposed into features when studying the mechanism of factual knowledge, and which method performs best. The candidate methods include: Sparse Autoencoders (SAE), Principal Component Analysis (PCA), Independent Component Analysis (ICA), and random directions (RD).
The hyperparameters and methodological details are provided in Appendix \ref{appendix:Details of SAE, PCA and ICA}. 
Both PCA and ICA perform the decomposition using the same amount of data used for training SAEs. 
\paragraph{Experiment settings}
The preliminary experiments focus on decomposing MLP activations to obtain features. We use pre-trained SAEs from Gemma Scope \cite{lieberum-etal-2024-gemma}. Given MLP activations $\mathbf{h}$ at layer $l$, features are obtained through the encoder function:
\begin{equation}
    \mathbf{f}(\mathbf{h}) := \sigma(\mathbf{W}_{enc}\mathbf{h} + \mathbf{b}_{enc})
\end{equation}
where $:=$ denotes definition, $\sigma$ is the JumpReLU activation, $\mathbf{W}_{enc}$ is the encoder weight matrix, and $\mathbf{b}_{enc}$ is the encoder bias vector. Each element $f_{l,p}(\mathbf{h})$ represents the activation of the feature at layer $l$ and position $p$.

\begin{figure}
    \centering
    \includegraphics[width=\linewidth]{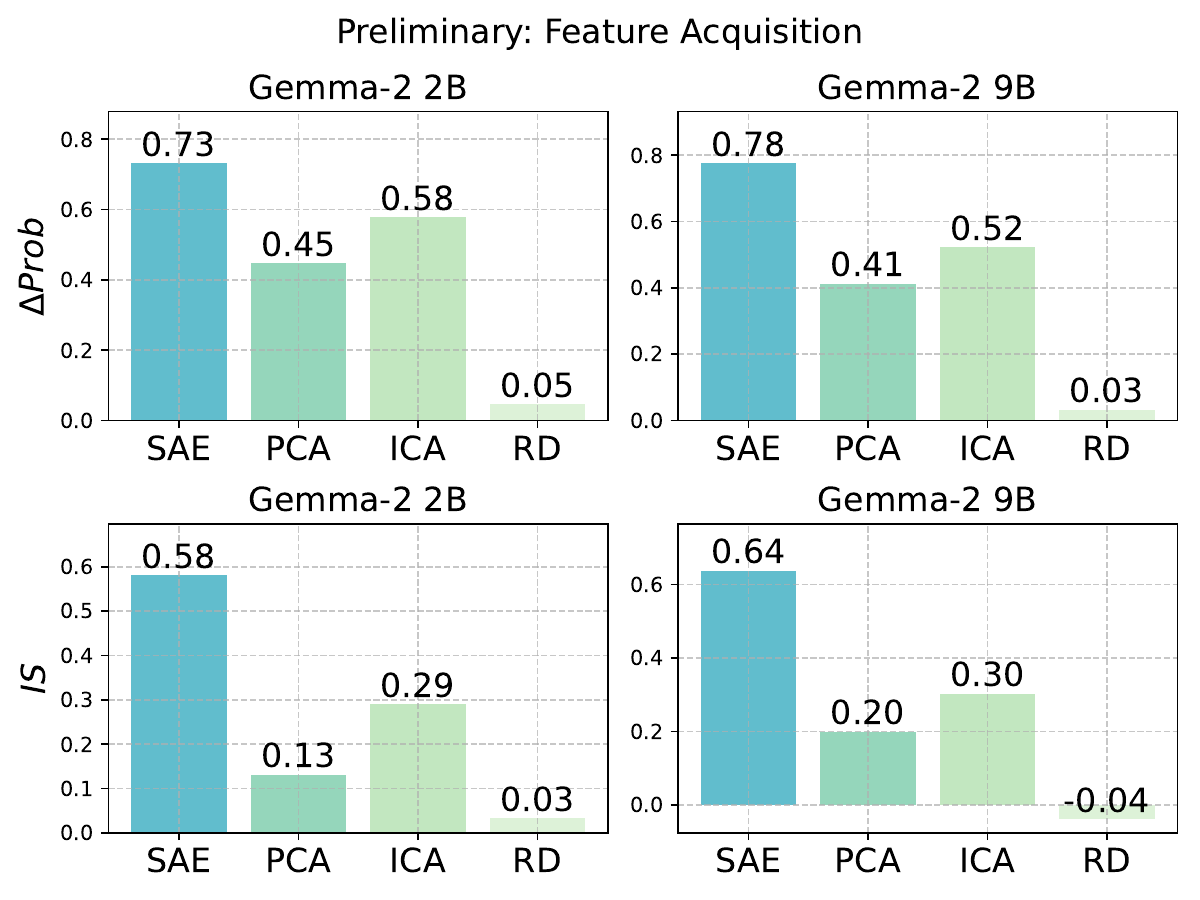}
\caption{Evaluation of features obtained by different methods. Top: $\Delta$ Prob after feature ablation. Bottom: Interpretation scores ($IS$). Higher values indicate better performance in both metrics.}
    \label{fig:Preliminary: Feature Acquisition:compare features}
\end{figure}

Structurally, these features in SAEs parallel the role of intermediate neurons in LLMs, as SAEs are trained to reconstruct LLMs in a higher-dimensional space. Taking one MLP layer as an example, both architectures follow a similar encoder-intermediate-decoder pattern:
\begin{align}
\label{eq:architecture_comparison}
    \text{LLMs}: & \mathbf{x} \xrightarrow{\text{encoder}} \text{intermediate neurons} \xrightarrow{\text{decoder}} \mathbf{y} \\
    \text{SAEs}: & \mathbf{h} \xrightarrow{\text{encoder}} \text{SAE  features} \xrightarrow{\text{decoder}} \mathbf{h}'
\end{align}
For a given input, we select highly activated features ($\mathbf{F_{a}}$) based on their spatial locations:
\begin{equation}
\label{equation:threshold1}
\mathbf{F_{a}} = \{(l,p) \mid f_{l,p}(\mathbf{h}) > \tau_1 \cdot \max_{l',p'} f_{l',p'}(\mathbf{h})\}
\end{equation}
where $\tau_1$ is the threshold parameter. Using the metrics ($\Delta Prob$ and $IS$) defined in \textsection\ref{section:Dataset, Models and Evaluation metrics}, we compare our SAE-based approach with baseline methods. The results are presented in Figure \ref{fig:Preliminary: Feature Acquisition:compare features}.

\begin{figure*}
    \centering
    \includegraphics[width=\linewidth]{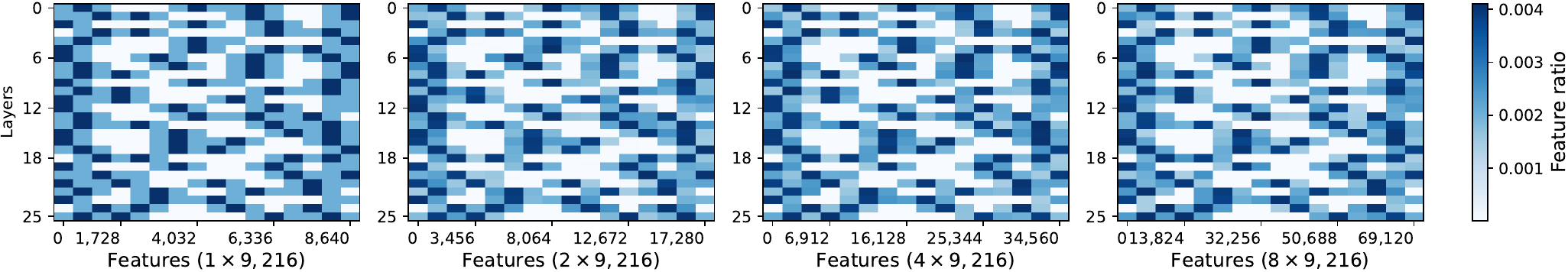}
\caption{Distribution plots of activated features under different feature number settings ($n\times 9216, n=1,2,4,8$) for Gemma-2 2B. 
The similar distribution patterns across different $n$ suggest that features consistently fall into similar regions. It should be noted that these four pictures are not exactly the same, but they are very similar.}
    \label{fig:Preliminary: Feature splitting}
\end{figure*}
\begin{figure*}
    \centering
    \includegraphics[width=\linewidth]{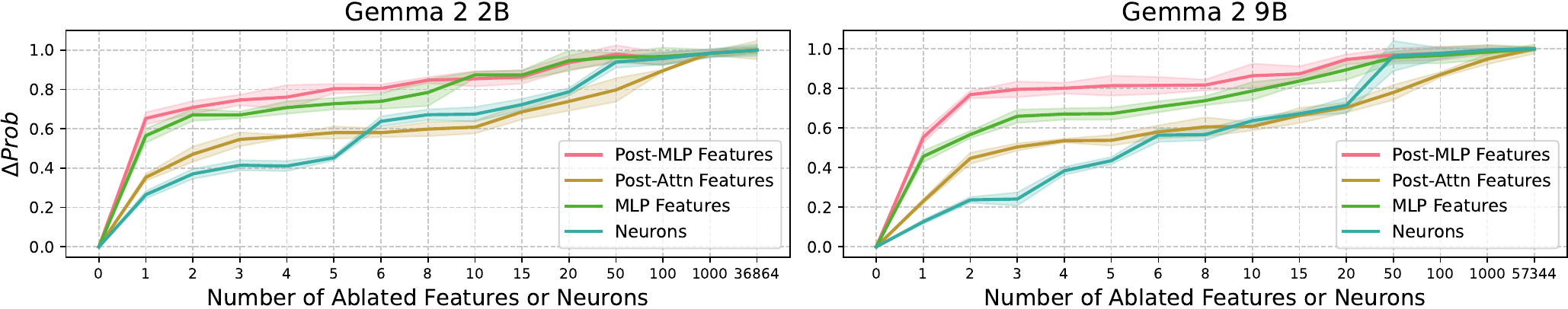}
\caption{The impact on $\Delta Prob$ when ablating features from different transformer components and neurons. Values show mean $\pm$ standard error across 5 bootstrap iterations, with higher values indicating greater influence on knowledge expression. Note that while $\Delta Prob \in [0,1]$, the plots may exceed 1 due to $+$ std.}
\label{fig:top2bottom_ablate_features_VS_neurons_both_models}
\end{figure*}

\paragraph{Findings}
\textbf{SAE demonstrates superior effectiveness in decomposing neurons into features compared to other methods.} For example, as shown in Figure \ref{fig:Preliminary: Feature Acquisition:compare features}, in Gemma-2 9b, SAE features demonstrate superior performance in both $\Delta Prob$ and $IS$, achieving $\Delta Prob$ of $\sim0.78$ and $IS$ of $\sim0.64$, showing increases of $\sim1.3\times$ in $\Delta Prob$ and $\sim2\times$ in $IS$ over the strongest baseline (ICA). Paired t-tests also confirm that SAE features are more effective for studying factual knowledge mechanisms (see Table \ref{table:appendix:statistical_tests:preliminary} in Appendix \ref{section:appendix:Paired T-test Results for Preliminary Experiment}).

\subsection{Feature Distribution Patterns Remain Consistent Across Feature Numbers}
When extracting features using SAE, we need to determine how many features to use for reconstructing LLM representations. As shown in Equation \ref{eq:architecture_comparison}, while LLMs use a fixed number of neurons, SAEs map these representations to a higher-dimensional feature space. The number of features ($N$) determines this dimensionality, typically set as $N = n \times d_{model}$ where $n$ is a multiplier and $d_{model}$ is the model's hidden dimension ($9216$ for Gemma-2 2B). Adjusting $N$ requires resource-intensive retraining. Notably, we observe consistent feature clustering patterns across different values of $N$.

In Figure \ref{fig:Preliminary: Feature splitting}, we visualize feature distributions across four different settings ($N = n \times 9216$) using Gemma-2 2B. The vertical axis represents layers, while the horizontal axis shows positions. The color intensity indicates the feature density in each bin, with darker blue representing higher values.  Using 500 randomly sampled facts\footnote{This sampling is necessary as visualizing features from the entire dataset would result in near-complete coverage of the feature space.}, we observe that \textbf{features consistently cluster in similar regions} when fixing the number of bins. As $N$ increases, these features undergo hierarchical decomposition within their original clusters, with larger $N$ values enabling finer-grained representations while maintaining the same overall distribution structure.

The results for Gemma-2 9B, along with a comprehensive quantitative analysis of the entire dataset presented in Appendix \ref{section:appendix-feature-stability} (Table \ref{appendix: tab:feature-stability} and Figure \ref{fig:appendix: 9B_feature_distribution}), corroborate these findings. One possible explanation could be that SAE features are insensitive to $N$. Based on this hypothesis, we fix $N=4 \times 9216$ for subsequent experiments, eliminating the need to compare different $N$ values in each experiment.

\section{Features vs. Neurons}
\label{section:Features vs. Neurons}
Building upon \textsection\ref{section:preliminary}, this section delves into the mechanism of factual knowledge using features obtained through sparse autoencoders (SAE). Our findings address the research questions Q1 (\textsection\ref{subsection:Post-MLP Features Have the Strongest Impact on Knowledge Expression} and \textsection\ref{subsection:Features Demonstrate Superior Interpretability Compared to Neurons}) and Q2 (\textsection\ref{subsection: Features Exhibit Stronger Monosemanticity}) raised in \textsection \ref{section:Introduction}.

\subsection{Post-MLP Features Have the Strongest Impact on Knowledge Expression}
\label{subsection:Post-MLP Features Have the Strongest Impact on Knowledge Expression}

\begin{figure}
    \centering
    \includegraphics[width=\linewidth]{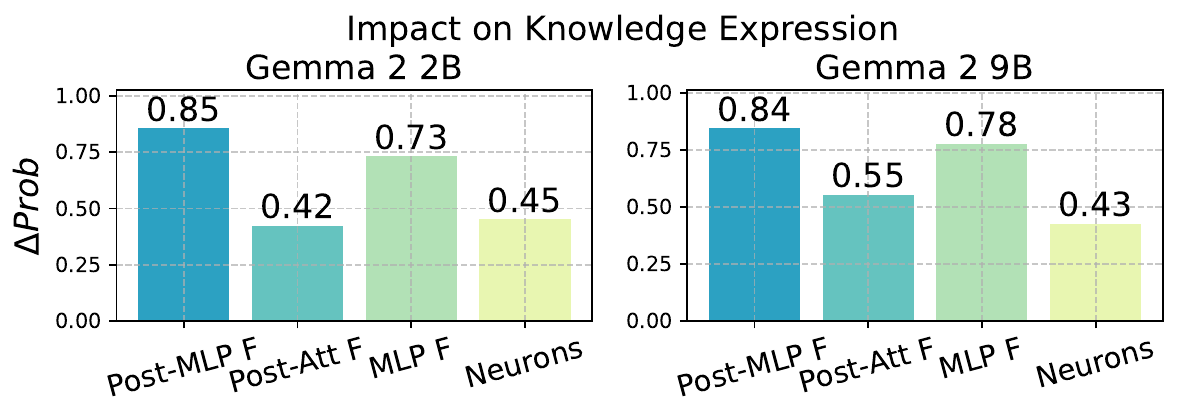}
\caption{The impact on $\Delta Prob$ when ablating features from different transformer components and neurons.}
    \label{fig:featureVSneurons_overall_Prob}
\end{figure}

\begin{figure*}
    \centering
    \includegraphics[width=\linewidth]{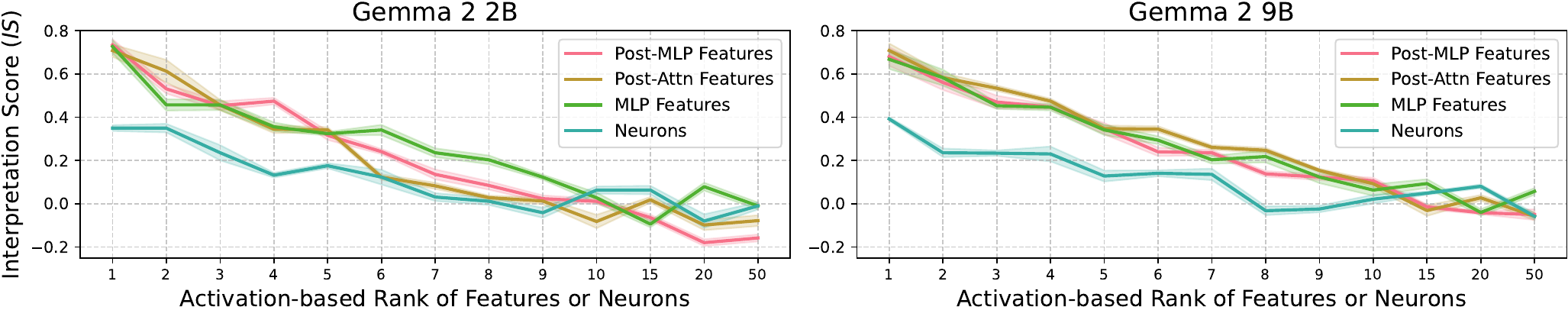}
\caption{Per-unit interpretation scores ($IS$) for features from different transformer components and neurons. We use the same bootstrap samples as Figure \ref{fig:top2bottom_ablate_features_VS_neurons_both_models}. Higher scores indicate better interpretability.}
    \label{fig:top2bottom_features_VS_neurons_IS}
\end{figure*}

\begin{figure}
    \centering
    \includegraphics[width=\linewidth]{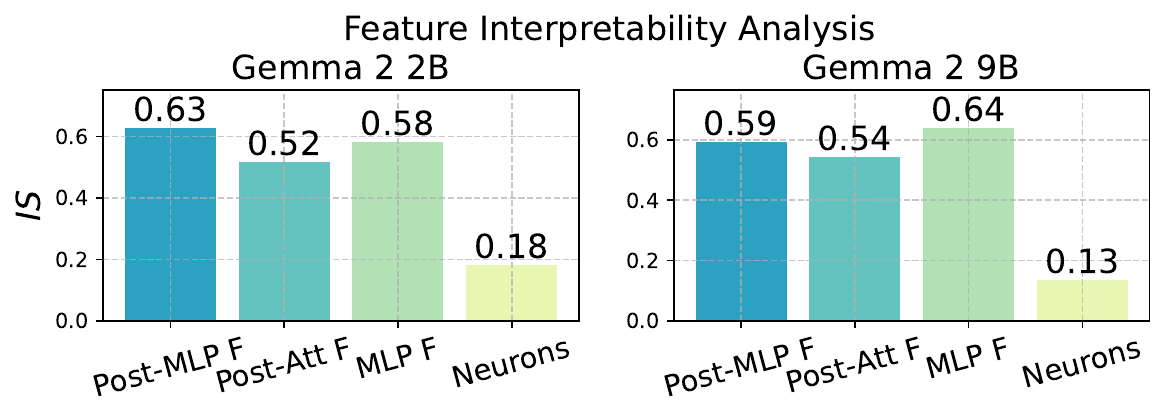}
\caption{The average interpretability scores ($IS$) for features from different components and neurons.}
\label{fig:featureVSneurons_overall_IS}
\end{figure}
\paragraph{Experiment Settings}
Following \textsection\ref{section:preliminary}, we extend our analysis to three components in transformer: post-attention residual, MLP activation, and post-MLP residual.
We apply SAE to extract features from each component and compare them with knowledge neurons identified using the localization method proposed by \citet{chen2024journey} (detailed in Appendix \ref{appendix:section:Knowledge Localization Method}), as their approach achieves state-of-the-art performance. We employ the $\Delta Prob$ metric (i.e., the decrease in model prediction probability) from \textsection\ref{section:Dataset, Models and Evaluation metrics} and conduct two complementary analyses to compare how features and neurons impact knowledge expression, with results shown in Figure \ref{fig:featureVSneurons_overall_Prob} and Figure \ref{fig:top2bottom_ablate_features_VS_neurons_both_models}.

(1) Figure \ref{fig:featureVSneurons_overall_Prob}: We select features and neurons through a thresholding method on the full dataset, then ablate them and calculate the $\Delta Prob$. 
For features, the selection follows Equation \ref{equation:threshold1}, and a similar thresholding technique with $\tau_1$ is applied for neurons.

(2) Figure \ref{fig:top2bottom_ablate_features_VS_neurons_both_models}: We perform a fine-grained analysis by ranking features based on their activations in descending order and progressively ablating them, calculating $\Delta Prob$ at each step. This approach allows us to observe the continuous impact of feature ablation without being constrained by any predetermined threshold. 
Since  progressive feature ablation on the full dataset is computationally intensive, we employ bootstrap sampling with 5 independent iterations (300 instances each, with replacement).

\paragraph{Findings}
\textbf{Post-MLP features have the strongest impact on knowledge expression.}
In Figure \ref{fig:featureVSneurons_overall_Prob}, the post-MLP features show a substantial impact with $\Delta Prob$ of $\sim0.85$, which is approximately $10\%$ higher than the strongest baseline (MLP features) and $\sim1.9\times$ that of neurons. Figure \ref{fig:top2bottom_ablate_features_VS_neurons_both_models} provides more granular evidence, showing that ablating just a few highly-activated features significantly impairs the model's ability to express knowledge. Ablating a single post-MLP feature yields a $\Delta Prob$ of $\sim0.6$, substantially higher than the strongest baseline (MLP features) and $\sim3\times$ that of neurons. The statistical significance test results are in Appendix \ref{subsection:appenidx:Paired T-test Results for prob and is} (Table \ref{appendix: tab:feature-stability}), confirming that the superior knowledge expression capabilities of post-MLP features over other features (or neurons) are significant.

\subsection{Features Demonstrate Superior Interpretability Compared to Neurons}

\label{subsection:Features Demonstrate Superior Interpretability Compared to Neurons}

\paragraph{Experiment Settings}
We employ the interpretability score ($IS$) metric introduced in \textsection\ref{section:Dataset, Models and Evaluation metrics} to evaluate the features and neurons, with results shown in Figure \ref{fig:featureVSneurons_overall_IS} and Figure \ref{fig:top2bottom_features_VS_neurons_IS}.

(1) Figure \ref{fig:featureVSneurons_overall_IS}: After obtaining features and neurons through the thresholding technique, we evaluate the interpretability scores ($IS$) of these selected units on the full dataset.

(2) Figure \ref{fig:top2bottom_features_VS_neurons_IS}: We first rank features in descending order based on their activations using the same 1000 sampled facts, then evaluate $IS$ for each feature/neuron individually. Unlike the batch ablation analysis in Figure \ref{fig:top2bottom_ablate_features_VS_neurons_both_models}, this approach assesses one unit at a time.
In this analysis, evaluating units individually holds greater significance beyond merely eliminating the influence of threshold. Since a single fact typically activates a larger number of neurons (can reach 20 or more) compared to features,  averaging $IS$ across neurons would bias the score toward lower values.  Therefore, evaluating each unit individually ensures a more equitable comparison. Note that we only evaluate up to the 50th unit, as the $IS$ approaches or falls below zero near this point, making further evaluation unnecessary.
\begin{figure*}
    \centering
    \includegraphics[width=\linewidth]{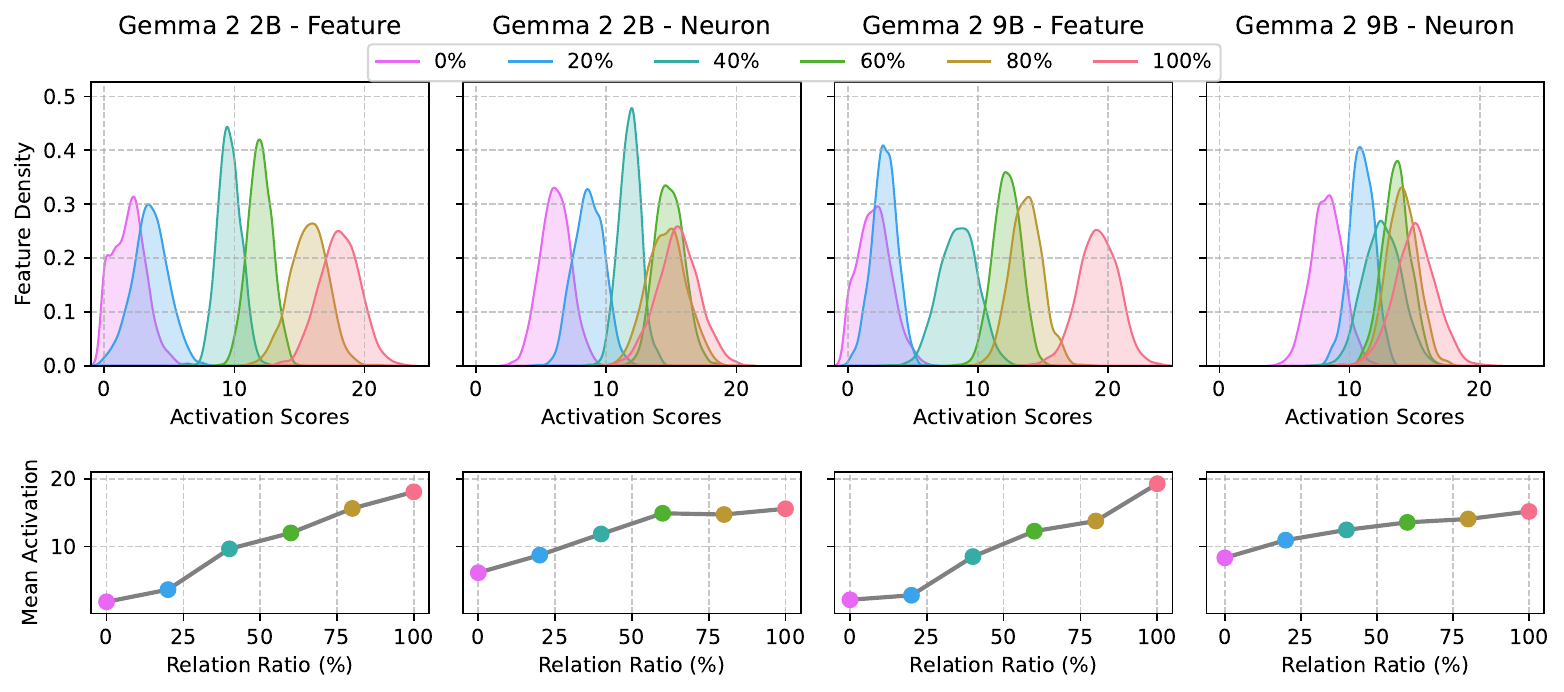}
\caption{Post-MLP feature activations corresponding to relation-facts under varying input compositions (0\% to 100\% relation-facts). Top: Activation score distributions. Bottom: Mean activation values.}
    \label{fig:features_facts_one_to_one}
\end{figure*}
\paragraph{Findings}
(1) \textbf{Features demonstrate superior interpretability compared to neurons.} In Figure \ref{fig:featureVSneurons_overall_IS}, post-MLP features achieve $IS$ values of $\sim0.6$, $\sim4\times$ that of neurons. A fine-grained analysis in Figure \ref{fig:top2bottom_features_VS_neurons_IS} further validates this conclusion, showing that even when compared individually, highly activated neurons consistently exhibit lower $IS$ values ($<0.4$) than highly activated features ($\sim0.7$).

(2) {Post-MLP features are a better choice when considering both metrics.} While MLP and post-attention features show comparable interpretability scores ($IS$ $\sim0.6$ and $\sim0.5$ respectively), post-MLP features consistently perform well in both interpretability and knowledge expression. The statistical significance test results are presented in Appendix \ref{subsection:appenidx:Paired T-test Results for prob and is} (Table \ref{appendix: tab:feature-stability}), showing that post-MLP features significantly outperform neurons in interpretability, and perform similarly to other features.

Let's review Q1: \textsection\ref{subsection:Post-MLP Features Have the Strongest Impact on Knowledge Expression} and \textsection\ref{subsection:Features Demonstrate Superior Interpretability Compared to Neurons} demonstrate that \textbf{features, as research units, effectively address the dual challenges of limited knowledge expression and poor interpretability.}

\subsection{Features Exhibit Stronger Monosemanticity}
\label{subsection: Features Exhibit Stronger Monosemanticity}
\paragraph{Motivation}
While our previous analyses demonstrate features' superior performance in both knowledge expression and interpretability, these findings indirectly suggest stronger monosemanticity of features. We now seek direct evidence to verify whether features better align with our desired scenario illustrated in Figure \ref{fig:introduction_figure1}(b). This would further support our findings in \textsection\ref{subsection:Post-MLP Features Have the Strongest Impact on Knowledge Expression} and \textsection\ref{subsection:Features Demonstrate Superior Interpretability Compared to Neurons}\footnote{Based on the analysis in \textsection\ref{subsection:Post-MLP Features Have the Strongest Impact on Knowledge Expression} and \textsection\ref{subsection:Features Demonstrate Superior Interpretability Compared to Neurons}, we focus our comparison specifically on post-MLP features and neurons.}.

\paragraph{Experiment Settings} To evaluate monosemanticity, we expect features corresponding to specific facts to show high activation values for those facts and low activation values for others. However, this evaluation presents two key challenges:

(1) Individual fact analysis is unreliable because each fact activates only a small subset of features. This sparsity means features corresponding to different facts might appear separated by chance, rather than due to true monosemanticity.
(2)  Analyzing all facts simultaneously would involve too many features, likely producing high activation values for some features regardless of input facts. This noise would mask the underlying feature-fact relationships we aim to study.

We address these challenges through a three-step approach:
(1) \textit{Relation Selection}: We select 5 relations (P39, P264, P37, P108, P131; see Appendix \ref{section:appendix-dataset}) and designate these facts as \textit{relation-facts}.

(2) \textit{Input Construction}: We maintain a constant total input size (2,591 facts) while varying the proportion of relation-facts from 0\% to 100\% in 20\% increments. For example, the ``40\%'' configuration contains 1,036 randomly sampled relation-facts and 1,555 non-relation facts.

(3) \textit{Activation Analysis}:
For each configuration, we:
(a) Record activation values from features (or neurons) associated with relation-facts.
(b) Visualize distributions using kernel density estimation (KDE), , where the $x$-axis is activation scores and the $y$-axis shows feature density.
(c) Plot mean activation values for clearer interpretation.

In this setup, stronger monosemanticity is characterized by clear distribution separation across different input configurations and higher activation scores when relation-fact proportions increase. Results are shown in Figure \ref{fig:features_facts_one_to_one}.

\paragraph{Findings}
\textbf{We resolve Q2: features exhibit superior monosemanticity compared to neurons.} 
(1) Features display distinct, well-separated activation waves that correlate with relation-fact proportions. In contrast, neurons show overlapping distributions and activation even without relevant inputs, i.e., $0\%$ condition (Top of Figure \ref{fig:features_facts_one_to_one}).

(2) Feature activation values increase systematically with relation-fact proportion, while neuron mean activation values start notably above zero and show minimal variation, especially between $60\%$ and $100\%$ (Bottom of Figure \ref{fig:features_facts_one_to_one}).

These findings demonstrate that features exhibit stronger fact-specific correspondence by remaining unresponsive to irrelevant inputs. Statistical analysis confirms this separation (all $p < 0.001$, Cohen's d$ > 0.8$; see Appendix \ref{subsection:appenidx:Paired T-test Results for Monosemanticity}).

\section{Feature-based Knowledge Erasure}

\label{section:Privacy Protection}
We evaluate our feature-based method against neuron-based approaches in erasing privacy-related information from LMs. This downstream application addresses Q3 raised in \textsection \ref{section:Introduction}, further validates our previous analysis, and demonstrates the practical value of our findings and analysis.

\subsection{Dataset} 
We construct \textbf{PrivacyParaRel}, a dataset containing synthetic privacy-sensitive information, following the triple format used by \citet{elazar2021measuring-dataset}.  Each entry is structured as \textit{$\langle$subject, relation, object$\rangle$}, such as \textit{$\langle$Alice, Social Security Number, 123-45-6789$\rangle$}, with multiple query variations generated for each fact. This format maintains consistency with factual knowledge datasets, enabling direct method transfer while addressing privacy concerns through synthetic data. We generate 1,500 different facts, each accompanied by six different query variations, resulting in 9,000 total entries. Further details in Appendix \ref{appendix:privacy-dataset}.

\subsection{Experiment Settings}

To erase specific knowledge from LMs, we first perform incremental fine-tuning on our privacy dataset, allowing the model to learn the private information. For erasure, we extract neurons and post-MLP features, then explore two approaches based on neurons and features respectively. In both approaches, we modify weights in the MLP layers to ensure fair comparison, as neuron-based methods operate on MLP weights. Note that this weight modification differs from the activation-zeroing approach used in $\Delta Prob$ calculation.

\paragraph{Neuron-based approach} Following existing neuron-based knowledge editing methods \citep{dai2022knowledge,chen2024knowledgelocalizationmissionaccomplished}, for each identified neuron $n^i_l$ in layer $l$, we set the $i$-th column to zero in $\mathbf{W}^{(2)}_l \in \mathbb{R}^{d_{io} \times d_m}$, where $\mathbf{W}^{(2)}_l$ is the second linear transformation matrix in the $l$-th MLP layer. Here, $d_{io}$ denotes the input/output dimension, and $d_m$ represents the intermediate dimension.

\paragraph{Feature-based approach}
We propose \textbf{FeatureEdit}, a reconstruction-based approach for feature-based model editing. Since features are not naturally exist in LMs, directly using them for model editing poses challenges.  To address this, \textbf{FeatureEdit} presents the first feature-based editing method, inspired by the activation reconstruction method used for $\Delta Prob$ (Appendix \ref{subsection:appendix:Feature Ablation Proces}).
For each identified feature $f^i_l$ in the $l$-th MLP layer, we create a one-hot probe vector ($\mathbf{p}^{i}_j$):
\begin{equation}
   \mathbf{p}^{i}_j = \begin{cases} 
   1 & \text{if} \quad j = i \\
   0 & \text{otherwise}
   \end{cases}
\end{equation}
Let $\mathbf{W}_e \in \mathbb{R}^{d_f \times d_m}$ be the encoding matrix learned by SAE. By reconstructing through its transpose (decoder matrix) $\mathbf{W}_e^T$, we obtain the feature's contribution pattern in the original MLP activation space:
\begin{equation}
   \mathbf{h}^{i} = \mathbf{W}_e^T\mathbf{p}^{i},\quad \mathbf{h}^{i} \in \mathbb{R}^{d_m}
\end{equation}
The reconstructed vector $\mathbf{h}^{i}$ reveals the feature's distributed influence. We traverse all features $f^i_l$, and identify significant positions in the weight matrix $\mathbf{W}^{(2)}_l$ by thresholding:
\begin{equation}
P = \{(l, c, i) \Big| |\mathbf{h}^{i}_c| > \tau_2, \text{ for all } i,l\}
\end{equation}
where $(l, c, i)$ represents the position in layer $l$, the $c$-th column and $i$-th row of $\mathbf{W}^{(2)}_l$, $\mathbf{h}^{(i)}_c$ is the $c$-th value of $\mathbf{h}^{(i)}$, and $\tau_2$ is a hyperparameter (Appendix \ref{subsection:appendix,Method-specific Parameters}). Finally, we zero out these specific weights: 
\begin{equation}
   \mathbf{W}^{(2)}_{l,c} = 0, \quad \forall (l,c,i) \in P
\end{equation}
Notably, this method achieves finer granularity than neuron-based approaches by enabling selective modification of specific weight positions rather than entire column vectors.

\paragraph{Evaluation Metrics} We employ four metrics to assess knowledge erasure performance \citep{yao2024knowledge,chen2024knowledgelocalizationmissionaccomplished}:
(1) Reliability (Rel): The probability that the model fails to correctly answer privacy-related queries after erasure.
(2) Generalization (Gen): The probability that the model fails to answer privacy-related queries with different phrasings. This metric is crucial as high Reliability with low Generalization indicates potential ``jailbreak'' \citep{NEURIPS2023_fd661313} phenomenon where models reveal private data in specific contexts.
(3) Locality (Loc): The probability that the model correctly answers unrelated queries. This metric ensures that knowledge erasure maintains other model capabilities.
(4) Perplexity: Measures the impact on the model's general text generation ability. We use $\Delta$PPL to quantify the perplexity change before (b) and after (a) erasure: $\Delta \text{PPL} = \frac{\text{PPL}_b-\text{PPL}_a}{\text{PPL}_b}$.

\begin{figure*}
    \centering
    \includegraphics[width=\linewidth]{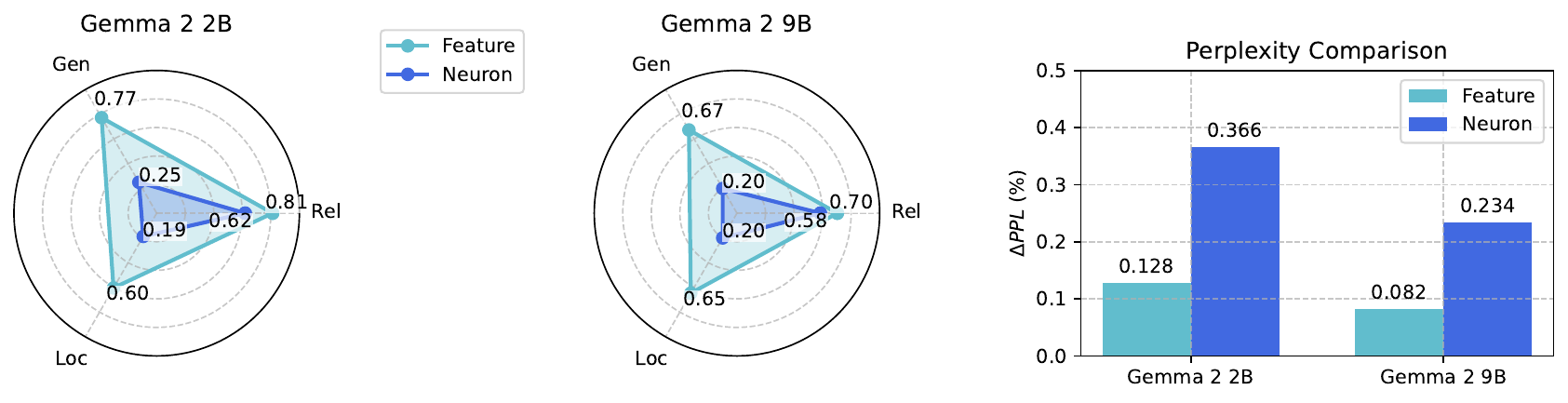}
    \caption{Results of knowledge erasure for privacy protection. For the radar chart metrics (Rel, Gen and Loc), higher values indicate better performance, while lower values in the bar chart indicate better performance.}
    \label{fig:privacy_erase}
    \vspace{-3mm}
\end{figure*}
\subsection{Findings}
\textbf{We resolve Q3: Feature-based model editing outperforms neuron-based methods in privacy knowledge erasure}. As shown in Figure \ref{fig:privacy_erase}, features achieve higher Rel scores ($\sim0.8$) compared to neurons ($\sim0.65$), indicating better erasure effectiveness. The substantially higher Gen score for features ($\sim0.7$ vs. $\sim0.25$) demonstrates significant mitigation of the ``jailbreak'' phenomenon. Moreover, features exhibit fewer side effects, evidenced by higher Loc scores ($\sim0.7$ vs. $\sim0.2$ for neurons), indicating minimal impact on other facts, and lower $\Delta$PPL ($\sim0.1$ vs. $\sim0.3$), suggesting better preservation of model generation capabilities. These findings align with our previous conclusions: features exert stronger influence on knowledge expression (higher Rel and Gen) while exhibiting superior monosemanticity, thus requiring fewer editing locations (better Loc and $\Delta$PPL).

\section{Related Works}
\paragraph{Knowledge Neurons Theory and its Limitations}
In studying the factual knowledge mechanisms of LMs, researchers often employ the knowledge neuron (KN) theory. Initially, \citet{geva-etal-2021-transformer} propose that MLP modules simulate key-value memories to store information, while \citet{dai2022knowledge} introduce the concept of knowledge neurons, suggesting that these neurons can store ``knowledge''. The success of KN-inspired model editing methods \citep{meng2022locating, meng2023massediting} further supports the plausibility of the KN theory. However, the KN theory has its limitations. \citet{niu2024what} argue that it oversimplifies the real situation, while \citet{hase2023does} suggest that the location of knowledge localization may not align with the location of greatest impact on knowledge expression. 
\citet{chen2024knowledgelocalizationmissionaccomplished} further challenge the fundamental assumptions of KN theory, suggesting facts are distributed across neurons and different queries about the same fact may activate different KNs.
Additionally, \citet{bricken2023monosemanticity} find that the activation of a single neuron can have different meanings in different contexts.
Limitations in KN-inspired knowledge editing methods have also been identified \citep{li2024unveiling, yao-etal-2023-editing, cohen2024evaluating, hoelscher2023detecting, pinter2023emptying, zhao2023unveiling}. These model editing methods may fail to edit successfully or impair the LMs' general capabilities, indirectly suggesting limitations with the KN thesis.
\paragraph{Decomposing Neurons into Features}
In the domain of general text processing, numerous studies have explored the properties of features. 
\citet{elhage2022superposition} demonstrate in toy neural networks, a layer of dimension N may linearly represent many more than N feature, showing that a large set of sparse features can be represented in a lower-dimensional space.
% This insight spurs efforts to apply dictionary learning to decode superposition.
\citet{bricken2023monosemanticity} advance the theory by decomposing neurons into more fine-grained features, arguing for their superiority as units of analysis. Using sparse autoencoders (SAE), they confirm that features correspond to patterns of neuron activations. Subsequently, researchers have improved the SAE method, proposing SAE variants with enhanced performance \citep{rajamanoharan2024improving,gao2024scaling}. Additionally, \citet{huben2024sparse} extend this approach to larger-scale LMs, while \citet{templeton2024scaling} discover highly abstract features that both respond to and cause abstract behaviors.
However, research on factual knowledge mechanisms has not yet adopted the features perspective. This paper focuses on transforming the unit of analysis, aiming to address some problems in the domain of factual knowledge.

% , possibly due to its complexity and the fact that features are not naturally present in the model. Furthermore, no algorithms have been developed to edit features for model editing purposes.This paper focuses on transforming the unit of analysis, aiming to address long-standing challenges in the domain of factual knowledge.
\vspace{-3mm}
\section{Conclusion}
We investigate the mechanism of factual knowledge in LMs and propose a shift from neuron-based to feature-based analysis. Drawing inspiration from feature extraction methods, we first validate the effectiveness of SAE in extracting features for factual knowledge. Based on this new analysis unit, we make several key findings: Features exhibit greater influence on knowledge expressing than neurons, offer enhanced interpretability, and demonstrate superior monosemanticity. Additionally, our feature-based approach demonstrates better performance in erasing privacy-sensitive knowledge from LMs.

\section{Limitations}
A significant challenge lies in the increased complexity of feature-based methods compared to neuron-based approaches, as features do not constitute natural units of analysis. While neurons are inherent components of LLM architecture, features require additional training of SAE for extraction. Consequently, although we can derive deeper insights using features, translating these insights into model performance improvements presents considerable challenges. For instance, while feature ablation is straightforward, such operations do not modify model parameters. Mapping external features back to model parameters may require an additional mapping layer to establish correspondence between features and weights. Our preliminary approach is based on reconstruction methods, consistent with activation ablation, but this is suboptimal. Therefore, mapping back to weights likely requires further algorithmic innovation.

Additionally, we observe an intriguing clustering phenomenon of SAE features that merits further investigation. While our current analysis confirms that features undergo progressive decomposition with increasing $N$ while maintaining their cluster structure, more fundamental questions remain unexplored. Specifically, we aim to investigate whether the SAE feature space possesses an inherent stable structure and whether it is truly insensitive to $N$. Such investigations could provide deeper insights into the nature of SAE feature representations.

\bibliography{custom}

\appendix

\section{Experimental Dataset Introduction}
\label{section:appendix-dataset}
In our experiments, we selected the ParaRel dataset \citet{elazar2021measuring-dataset}, a high-quality resource of cloze-style query English paraphrases. It contains a total of 328 paraphrases for 38 relations. We further conducted a basic filtering, excluding 2 relations that had no paraphrases. Table \ref{appendix:tab:relation_examples} displays these relations and corresponding example data.
\begin{table*}[t!]
% \small
\centering
\begin{tabular}{l l l} 
\toprule
\multicolumn{1}{c}{\multirow{3}{*}{{\textbf{Relation}}}} & \multicolumn{2}{c}{{\textbf{Example data}}} \\
\cmidrule(lr){2-3}
& {\textbf{Example Query}} & {\textbf{Answer}} \\ 
\cmidrule(lr){2-2} \cmidrule(lr){3-3}
\cmidrule(lr){1-1}
{P39}   & Adrian IV has the position of & pope \\
{P264}  & Purple Hearts is represented by music label & Sunshine \\
{P37}   & The official language of Republic of Ingushetia is & Russian \\
{P108}  & Henry Swanzy works for & BBC \\
{P131}  & Heaton Park is located in & Manchester \\
{P103}  & The native language of Francis Ponge is & French \\
{P176}  & Fiat Grande Punto is produced by & Fiat \\
{P30}   & Somalia is located in & Africa \\
{P178}  & Gain Ground is developed by & Sega \\
{P138}  & International Day for Biological Diversity is named after & biodiversity \\
{P47}   & Ukraine shares border with & Poland \\
{P17}   & Media Development Authority is located in & Singapore \\
{P413}  & Joe Torre plays in [MASK] position. & catcher \\
{P27}   & Edward Wollstonecraft is [MASK] citizen. & Australia \\
{P463}  & Chuck Schuldiner is a member of & Death \\
{P364}  & The original language of NU.nl is & Dutch \\
{P495}  & The Creepshow was created in & Canada \\
{P449}  & Yes Minister was originally aired on & BBC \\
{P20}   & Margaret Cavendish, Duchess of Newcastle-upon-Tyne died in & England \\
{P1376} & Rumbek is the capital of & Lakes \\
{P1001} & Minister for Foreign Affairs is a legal term in & Australia \\
{P361}  & propellant is part of & cartridge \\
{P36}   & The capital of Flanders is & Brussels \\
{P1303} & Ludovico Einaudi plays & piano \\
{P530}  & Brunei maintains diplomatic relations with & Australia \\
{P19}   & Lopo Soares de Albergaria was born in & Lisbon \\
{P190}  & Bratislava and [MASK] are twin cities. & Dublin \\
{P740}  & Shirehorses was founded in & Manchester \\
{P136}  & Frank Mantooth plays [MASK] music. & jazz \\
{P127}  & AVCHD is owned by & Sony \\
{P1412} & Karl Bodmer used to communicate in & French \\
{P407}  & Zarez was written in & Croatian \\
{P140}  & Leo IX is affiliated with the [MASK] religion. & Christianity \\
{P279}  & quinquina is a subclass of & wine \\
{P276}  & Al-Rifa'i Mosque is located in & Cairo \\
{P159}  & The headquarter of Allied Command Transformation is in & Norfolk \\
{P106}  & Giuseppe Saracco is a [MASK] by profession. & politician \\
{P101}  & Aleksei $N$. Leontiev works in the field of & psychology \\
{P937}  & Joseph Chamberlain used to work in & London \\
\bottomrule
\end{tabular}
\caption{Example data of the ParaRel dataset \citep{elazar2021measuring-dataset}.}
\label{appendix:tab:relation_examples}
\end{table*}

\section{Experimental Hardware Specification and Environment}
All experiments are conducted using a high-performance computing system with an Intel(R) Xeon(R) CPU E5-2680 v4 (2.40GHz, 56 cores) processor and 10 NVIDIA GeForce RTX 3090 GPUs, each equipped with 24576 MiB of memory. The software environment consists of Python 3.10.10 and PyTorch 2.0.0+cu117 for deep learning implementations.

\section{Feature Ablation Process and Autointerpretation Protocol}
\label{section:appendix:Autointerpretation Protocol}

Here we will introduce in detail how we obtain $\Delta Prob$ and $IS$.
\subsection{Feature Ablation Process}

\label{subsection:appendix:Feature Ablation Proces}
Let $\mathbf{h} \in \mathbb{R}^{d_m}$ denote the original component activation (e.g., MLP activation) at a specific layer. Through SAE, we obtain the encoding matrix $\mathbf{W}_e \in \mathbb{R}^{d_f \times d_m}$ and feature vector $\mathbf{f} = \sigma(\mathbf{W}_e\mathbf{h}) \in \mathbb{R}^{d_f}$, where $d_f$ is the number of features and $\sigma$ is the activation function.
The feature ablation process follows these steps:

1. Given a set of target features to ablate $S$, we create a masked feature vector $\mathbf{f}'$:
\begin{equation}
   \mathbf{f}'_i = \begin{cases} 
   0 & \text{if } i \in S \\
   \mathbf{f}_i & \text{otherwise}
   \end{cases}
\end{equation}

2. We reconstruct the activation using the decoder matrix $\mathbf{W}^T_e$:
   \begin{equation}
   \mathbf{h}' = \mathbf{W}^T_e\mathbf{f}' \in \mathbb{R}^{d_m}
\end{equation}

3. Replace the original activation $\mathbf{h}$ with the reconstructed activation $\mathbf{h}'$ in the model's forward computation to obtain the modified probability $Prob_a$.
This process allows us to measure how specific features influence the model's knowledge expression by comparing the original probability $Prob_b$ (using $\mathbf{h}$) with the modified probability $Prob_a$ (using $\mathbf{h}'$) through the $\Delta Prob$ metric.

\subsection{Autointerpretation Protocol}
\label{subsection:appendix:Autointerpretation Protocol}
We adapt the interpretability evaluation method from \citet{bills2023language} for our factual knowledge dataset, which consists of triples in various domains. This method is applied to features extracted by Sparse Autoencoders (SAE) from LLMs' post-MLP residual flow (this paper uses Gemma 2 2B and Gemma 2 9B). The process for each feature is as follows:

1. We select 20 diverse samples from our dataset of factual knowledge triples. Each sample is run through LLMs, measuring the feature's activation (range 0-1).

2. We identify the top 3 samples with highest feature activation. These high-activation samples are provided to a large language model. (we use gpt-4o-mini here\footnote{Any large language model can be used, but it is required that this LLMs can output logprobs.}.)

3. Based on this interpretation, we ask gpt-4o-mini to predict activation levels for 6 new samples: 3 high-activation and 3 random samples from our dataset.

4. We calculate the correlation between these predictions and the actual Gemma 2 2B activations, yielding an interpretability score for the feature.
\section{Details of SAE, PCA, ICA and FeatureEdit}
\label{appendix:Details of SAE, PCA and ICA}

This section details the four methods used for extracting interpretable features: Sparse Autoencoders (SAE), Principal Component Analysis (PCA), Independent Component Analysis (ICA), and random directions. Assume that the input is MLP activation. Other inputs are similar.

\subsection{JumpReLU Sparse Autoencoders (SAEs)}
JumpReLU SAEs are neural networks that learn sparse representations through a threshold-based activation mechanism \citep{lieberum-etal-2024-gemma}. Given MLP activations $\mathbf{h} \in \mathbb{R}^{d_m}$, the encoder and decoder functions are defined by:
\begin{equation}
    \mathbf{f}(\mathbf{h}) := \sigma(W_{enc}\mathbf{h} + \mathbf{b}_{enc})
\end{equation}
\begin{equation}
    \hat{\mathbf{h}}(\mathbf{f}) := W_{dec}\mathbf{f} + \mathbf{b}_{dec}
\end{equation}
where $W_{enc} \in \mathbb{R}^{d_f \times d_m}$, $W_{dec} \in \mathbb{R}^{d_m \times d_f}$, $\mathbf{b}_{enc} \in \mathbb{R}^{d_f}$, $\mathbf{b}_{dec} \in \mathbb{R}^{d_m}$.

The JumpReLU activation $\sigma$ is defined as:
\begin{equation}
    \sigma(\mathbf{z}) = \text{JumpReLU}_{\theta}(\mathbf{z}) := \mathbf{z} \odot H(\mathbf{z} - \theta)
\end{equation}
where $\theta > 0$ is the learnable threshold parameter and $H$ is the Heaviside step function.

The loss function combines reconstruction error with an L0 sparsity penalty:
\begin{equation}
    \mathcal{L} = \|\mathbf{h} - \hat{\mathbf{h}}(\mathbf{f}(\mathbf{h}))\|_2^2 + \lambda\|\mathbf{f}(\mathbf{h})\|_0
\end{equation}
where $\lambda$ controls the sparsity penalty weight.

Features are obtained through:
\begin{equation}
    \mathbf{F} = \{\mathbf{f}(\mathbf{h}) := \sigma(W_{enc}\mathbf{h} + \mathbf{b}_{enc})\}
\end{equation}

Selected features are identified using:
\begin{equation}
\label{identifyFa}
    \mathbf{F_{a}} = \{f \in \mathbf{F} \mid a(f) > \tau_1 \cdot \max_{f \in \mathbf{F}} a(f)\}
\end{equation}
In this equation, $a(f)$ represents the activation value of feature $f_i$, while $\tau_1$ serves as the threshold parameter controlling feature selection sensitivity. The term $\max_{f \in \mathbf{F}} a(f)$ denotes the maximum activation value across all features.

\subsection{Principal Component Analysis (PCA)}
PCA finds orthogonal directions that capture maximum variance in the data. For MLP activations $\mathbf{H} = [\mathbf{h}_1, ..., \mathbf{h}_{d_m}]^T$, the process involves several key steps. First, we center the data by computing $\mathbf{H}_c = \mathbf{H} - \mathbb{E}[\mathbf{H}]$. Next, we compute the covariance matrix $\mathbf{C} = \frac{1}{n}\mathbf{H}_c^T\mathbf{H}_c$. We then perform eigendecomposition $\mathbf{C} = \mathbf{V}\Lambda\mathbf{V}^T$, where $\mathbf{V} = [\mathbf{v}_1, ..., \mathbf{v}_{d_m}]$ contains eigenvectors. Finally, we project the data using $\mathbf{F} = \mathbf{H}_c\mathbf{V}_{d_f}$, where $\mathbf{V}_{d_f}$ contains top $d_f$ eigenvectors.

Features are obtained through:
\begin{equation}
    \mathbf{F} = \{\mathbf{f}(\mathbf{h}) := \mathbf{h}^T\mathbf{V}_{d_f}\}
\end{equation}

Selected features are identified using Equation \ref{identifyFa}.
\subsection{Independent Component Analysis (ICA)}
ICA seeks to find statistically independent components by maximizing non-Gaussianity. The process begins with whitening, where we transform the data to have unit variance in all directions:
\begin{equation}
    \mathbf{H}_w = \mathbf{H}_c\mathbf{V}\Lambda^{-1/2}
\end{equation}
We then find the unmixing matrix $\mathbf{W} \in \mathbb{R}^{d_f \times d_m}$ that maximizes non-Gaussianity:
\begin{equation}
    \mathbf{F} = \mathbf{H}_w\mathbf{W}
\end{equation}
The optimization typically uses approximations of negentropy:
\begin{equation}
    J(\mathbf{w}) = [E\{G(\mathbf{w}^T\mathbf{h}_w)\} - E\{G(\nu)\}]^2
\end{equation}
where $G$ is a non-quadratic function and $\nu$ is a standard Gaussian variable.

Features are obtained through:
\begin{equation}
    \mathbf{F} = \{\mathbf{f}(\mathbf{h}) := \mathbf{h}_w^T\mathbf{W}\}
\end{equation}

Selected features are identified using Equation \ref{identifyFa}.

\subsection{Random Directions (RD)}
Random directions serve as a baseline method through a three-step process. Initially, we generate a random matrix $\mathbf{R} \in \mathbb{R}^{d_m \times d_f}$ with entries drawn from $\mathcal{N}(0, 1/\sqrt{d_m})$. We then apply QR decomposition to obtain an orthonormal basis:
\begin{equation}
    \mathbf{R} = \mathbf{Q}\mathbf{R}_{upper}
\end{equation}
where $\mathbf{Q} \in \mathbb{R}^{d_m \times d_f}$ is an orthonormal matrix and $\mathbf{R}_{upper} \in \mathbb{R}^{d_f \times d_f}$ is an upper triangular matrix. Finally, we project the data using the orthonormal matrix: $\mathbf{F} = \mathbf{H}\mathbf{Q}$.

Features are obtained through:
\begin{equation}
    \mathbf{F} = \{\mathbf{f}(\mathbf{h}) := \mathbf{h}^T\mathbf{Q}\}
\end{equation}

Selected features are identified using Equation \ref{identifyFa}.

\begin{figure*}
    \centering
    \includegraphics[width=\linewidth]{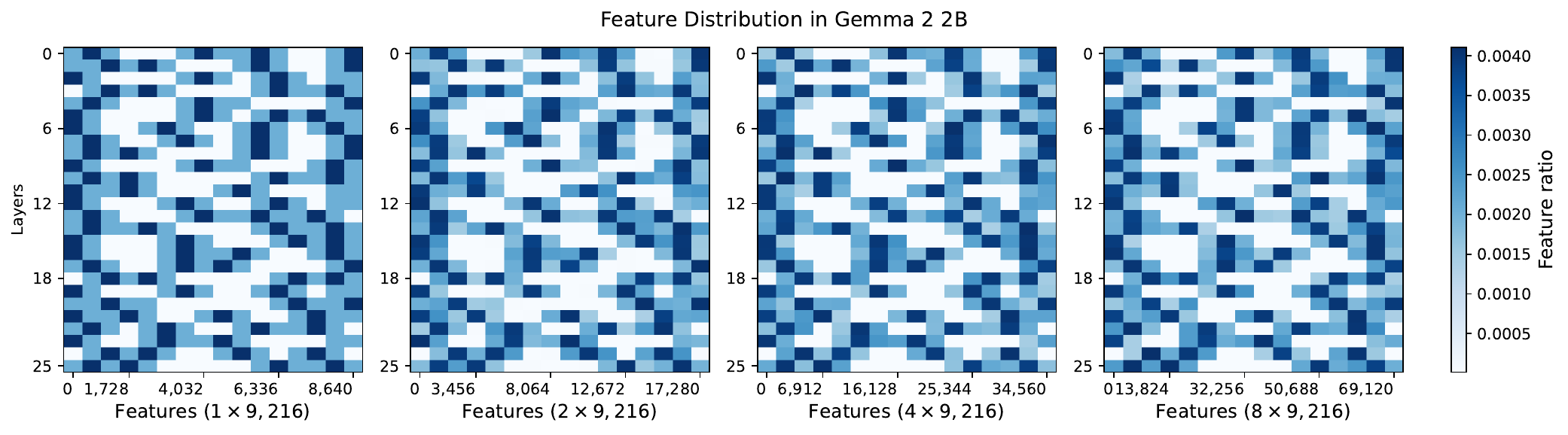}
\caption{Feature cluster Results for Gemma 2 2B.}
    \label{fig:appendix: 2B_feature_distribution}
\end{figure*}
\begin{figure*}
    \centering
    \includegraphics[width=\linewidth]{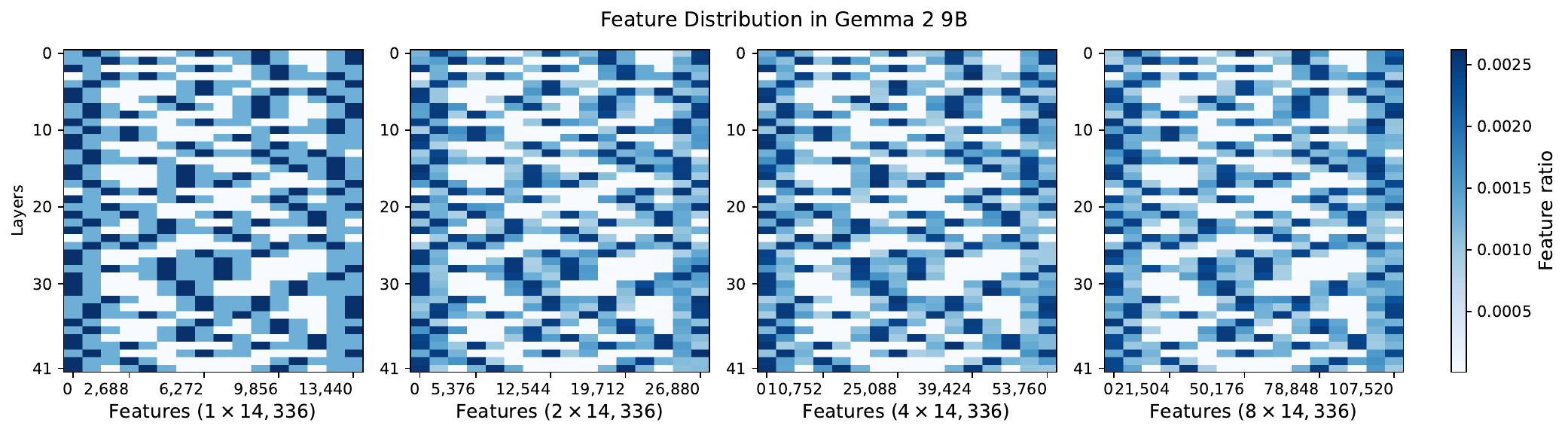}
\caption{Feature cluster Results for Gemma 2 9B.}
    \label{fig:appendix: 9B_feature_distribution}
\end{figure*}

\subsection{Method-specific Parameters}
\label{subsection:appendix,Method-specific Parameters}
The implementation of each method involves specific parameter settings. For SAE, we use $\beta = 3$, $\rho = 0.05$, sigmoid activation, and feature dimension $d_f = n \times d_m$, with $n=4$ in this paper. The training process employs the Adam optimizer with learning rate $1e^{-3}$, batch size 256, and runs for 100 epochs. Early stopping is triggered if validation loss does not improve for 10 consecutive epochs. Input activations are standardized to zero mean and unit variance before training.

PCA employs an explained variance ratio threshold of 0.95, which determines the resulting $d_f$ features. The input data is centered but not scaled, as variance information is crucial for principal component identification. ICA utilizes the FastICA algorithm with cubic $G$ function and feature dimension $d_f = 4d_m$, with input data whitened during preprocessing. The Random method uses Gaussian initialization with variance scaling and maintains a feature dimension of $d_f = 4d_m$.

For feature selection across all methods, we employ a threshold $\tau_1 = 0.3$ to identify significant features, striking a balance between feature coverage and selectivity. This threshold was determined through preliminary experiments examining the distribution of feature activations across different knowledge categories. Specifically, $\tau_1 = 0.3$ ensures capture of features that demonstrate substantial activation (at least 30\% of maximum activation) while filtering out noise and weakly activated features.

For FeatureEdit, we set the reconstruction threshold $\tau_2 = 0.1$ to identify significant weight positions. This threshold was chosen based on the empirical observation of weight contribution distributions in the reconstructed activation space, ensuring that we capture meaningful feature influences while maintaining editing precision. The relatively small threshold value allows us to identify subtle but important feature contributions in the distributed representations.

\section{Paired T-test Results for Preliminary Experiment}
\label{section:appendix:Paired T-test Results for Preliminary Experiment}
To rigorously validate the superiority of SAE features over baseline methods, we conduct paired t-tests using full dataset for each method. For both models (Gemma 2 2B and Gemma 2 9B) and both metrics ($\Delta Prob$ and $IS$), we compare SAE with each baseline method (PCA, ICA, and random baseline). The statistical significance of the differences is assessed using paired t-tests, as we compare different methods on the same set of instances. Table \ref{table:appendix:statistical_tests:preliminary} presents the detailed statistical analysis results.

Notably, we include Cohen's d effect size alongside traditional significance testing because p-values alone may not reflect the practical significance of the differences, especially with large sample sizes (all p<0.001). Cohen's d measures the standardized difference between two means, where values above 0.8 indicate large effects. Our results show substantial effect sizes (Cohen's d ranging from 0.38 to 3.81, with most values exceeding 0.8), confirming not only the statistical significance but also the practical importance of SAE's improvements over baseline methods. Particularly strong effects are observed when comparing SAE with the random baseline (Cohen's d > 1.6), and in the $IS$ metric where almost all comparisons show large effect sizes (Cohen's d > 0.7).

\begin{table*}[h]
\centering
\begin{tabular}{llccc}
\toprule
\multicolumn{5}{c}{\textbf{$\Delta Prob$}} \\
\toprule
\textbf{Model} & \textbf{Method} & \textbf{t-statistic} & \textbf{p-value} & \textbf{Cohen's d} \\
\midrule
Gemma 2 2B & SAE vs. PCA & 85.15 & $<0.001$ & 0.70 \\
Gemma 2 2B & SAE vs. ICA & 46.33 & $<0.001$ & 0.38 \\
Gemma 2 2B & SAE vs. Random & 359.48 & $<0.001$ & 2.94 \\
\midrule
Gemma 2 9B & SAE vs. PCA & 120.76 & $<0.001$ & 0.99 \\
Gemma 2 9B & SAE vs. ICA & 77.18 & $<0.001$ & 0.63 \\
Gemma 2 9B & SAE vs. Random & 466.02 & $<0.001$ & 3.81 \\
\midrule
\multicolumn{5}{c}{\textbf{$IS$}} \\
\toprule
\textbf{Model} & \textbf{Method} & \textbf{t-statistic} & \textbf{p-value} & \textbf{Cohen's d} \\
\midrule
Gemma 2 2B & SAE vs. PCA & 157.64 & $<0.001$ & 1.29 \\
Gemma 2 2B & SAE vs. ICA & 87.19 & $<0.001$ & 0.71 \\
Gemma 2 2B & SAE vs. Random & 200.49 & $<0.001$ & 1.64 \\
\midrule
Gemma 2 9B & SAE vs. PCA & 155.98 & $<0.001$ & 1.27 \\
Gemma 2 9B & SAE vs. ICA & 104.44 & $<0.001$ & 0.85 \\
Gemma 2 9B & SAE vs. Random & 255.96 & $<0.001$ & 2.09 \\
\bottomrule
\end{tabular}
\caption{Statistical analysis of feature acquisition methods. We report t-statistics, p-values from paired t-tests, and Cohen's d effect sizes for comparing SAE with baseline methods (PCA, ICA, and random baseline) across both metrics ($\Delta Prob$ and $IS$).}
\label{table:appendix:statistical_tests:preliminary}
\end{table*}

\section{Quantitative Analysis of Feature Stability Across Different $N$ Values}
\label{section:appendix-feature-stability}

To further validate our observation that the impact of the number of features ($N$) is less significant than anticipated, we conduct a comprehensive quantitative analysis on the entire dataset. This analysis aims to support our conclusion that our findings are stable across different values of the hyperparameter $N$.

\subsection{Methodology}

We define $N$ as $N = n \times \text{len(MLP activation)}$, where $n$ is a positive integer. We use $n = 1$ as the baseline for comparison. This approach yields $\text{layer} \times N$ features for each model.

Using Gemma 2 2B as an example, our methodology is as follows:

1. For a given fact, when $n = 1$, we record the positions of activated features as $[\text{layer}, \text{position}]$.

2. For any integer $n > 1$, based on the $n = 1$ case, we expect features to fall within the range $[\text{layer}, \text{position} \times n, (\text{position} + 1) \times n - 1]$.

3. We then compare the actual positions of features for $n > 1$ with these expected positions and calculate the overlap ratio.

4. We repeat this process for the entire dataset and compute the average overlap ratio.

We apply this methodology to both Gemma 2 2B and Gemma 2 9B models, using $n$ values of 1, 2, 4, and 8.

\subsection{Results}
Table \ref{appendix: tab:feature-stability} presents the average overlap ratios for different $n$ values across both models. Additionally, Figure \ref{fig:appendix: 9B_feature_distribution} complements the results shown in Figure \ref{fig:Preliminary: Feature splitting} from the main text. While Figure \ref{fig:Preliminary: Feature splitting} only presents the results for Gemma 2 2B, Figure \ref{fig:appendix: 9B_feature_distribution} displays the results for both Gemma 2 2B and Gemma 2 9B.

The results in Table \ref{appendix: tab:feature-stability} demonstrate a high degree of overlap between the expected and actual feature positions across different $n$ values. For both Gemma 2 2B and Gemma 2 9B, we observe that even as $n$ increases to 8, the overlap ratio remains above 0.87, indicating a strong consistency in feature localization.

\begin{table*}[htbp]
\centering
\begin{tabular}{@{}lcccc@{}}
\toprule
\multirow{2}{*}{Model} & \multicolumn{4}{c}{$n$} \\
\cmidrule(l){2-5}
& 1 & 2 & 4 & 8 \\
\midrule
Gemma 2 2B    & 1.000 & 0.927 & 0.891 & 0.893 \\
Gemma 2 9B & 1.000 & 0.935 & 0.908 & 0.879 \\
\bottomrule
\end{tabular}
\caption{Average overlap ratios for different $n$ values.}
\label{appendix: tab:feature-stability}
\end{table*}
This quantitative analysis supports our earlier observation that as $N$ increases, the original features are further decomposed but remain aggregated in consistent regions. The high overlap ratios suggest that our conclusions about feature behavior and importance are indeed stable and relatively insensitive to changes in the hyperparameter $N$.

These findings have important implications for future research in this area, as they suggest that the choice of $N$, within a reasonable range, does not significantly alter the fundamental patterns of feature activation and localization in relation to factual knowledge representation in language models.

\section{Knowledge Localization Method}

\label{appendix:section:Knowledge Localization Method}
We compare the precision of knowledge neuron localization across different research papers and select Architecture-adapted Multilingual Integrated Gradients \citep{chen2024journey} as our baseline method, as it demonstrates superior performance in knowledge neuron localization.

Given a query \(q\), they  define the probability of the correct answer predicted by a PLMs as follows:
\begin{equation}
\label{eq:1}
    \operatorname{F}(\hat{w}^{(l)}_{j}) = p(y^* | q, w^{(l)}_{j}=\hat{w}^{(l)}_{j})
\end{equation}
Here, \(y^*\) represents the correct answer, \(w^{(l)}_{j}\) denotes the \(j\)-th neuron in the \(l\)-th layer, and \(\hat{w}^{(l)}_{j}\) is the specific value assigned to \(w^{(l)}_{j}\). To calculate the attribution score for each neuron, they employ the technique of integrated gradients.
To compute the attribution score of a neuron \(w^{(l)}_{j}\), they consider the following formulation:
\begin{equation}
    \Delta w^{(l)}_{j} = \overline{w}^{(l)}_{j} - {w'}^{(l)}_{j}
\end{equation}
\begin{equation}
    \label{eqution:attribute}
     \operatorname{Attr}({w}^{(l)}_{j}) = \Delta w^{(l)}_{j} \int_{0}^{1} \frac{\partial \operatorname{F}({w'}^{(l)}_{j} + \alpha\Delta w^{(l)}_{j})}{\partial {w}^{(l)}_{j}}  \, d\alpha
\end{equation}Here, \(\overline{w}^{(l)}_{j}\) represents the actual value of \(w^{(l)}_{j}\),
\(w'^{(l)}_{j}\) serves as the baseline vector for \(w^{(l)}_{j}\). The term \(\frac{\partial \operatorname{F}(w'^{(l)}_{j} + \alpha\Delta w^{(l)}_{j})}{\partial w^{(l)}_{j}}\) computes the gradient with respect to \(w^{(l)}_{j}\).  
Next, they aim to obtain ${w'}^{(l)}_{j}$. Starting from the sentence $q$, they acquire a baseline sentence and then encode this sentence as a vector.
Let the baseline sentence corresponding to $q_i$ be $q'_i$, and $q'_i$ consists of $m$ words, maintaining a length consistent with $q$, denoted as $q'_i=(q'_{i1} \ldots q'_{ik} \ldots q'_{im})$. Since they are using auto-regressive models, according to \citet{chen2024journey}, $q'_{ik}=\langle \text{eos}\rangle$, where $\langle \text{eos}\rangle$ represents ``end of sequence'' in auto-regressive models.
The attribution score \(Attr_i(w_j^{(l)})\) for each neuron, given the input \(q_i\), can be determined using Equation \eqref{eqution:attribute}. For the computation of the integral, the Riemann approximation method is employed:
\begin{equation}
    {Attr_i(w_j^l)} \approx \frac{\overline{w}^{(l)}_{j}}{N} \sum_{k=1}^{N} \frac{ \partial F({w'}^{(l)}_{j} + \frac{k}{N} \times \Delta w^{(l)}_{j}}{\partial {w}^{(l)}_{j}}
\end{equation}where $N$ is the number of approximation steps.
Then, the attribution scores for each word \(q_i\) are aggregated and subsequently normalized:
\begin{equation}
    Attr(w_j^l) = \frac{\sum_{i=1}^{m} Attr_i(w_j^l)}{\sum_{j=1}^{n} \sum_{i=1}^{m} Attr_i(w_j^l)}
\end{equation}

Let \( \mathcal{N} \) be the set of neurons classified as knowledge neurons based on their attribution scores exceeding a predetermined threshold \( \tau \), for a given input \( q \). This can be formally defined as:

\begin{equation}
\mathcal{N} = \left\{ w_j^{(l)} \,\middle|\, Attr(w_j^{(l)}) > \tau \right\}
\end{equation}where \(l\) encompassing all layers and \(j\) including all neurons within each layer.

\begin{table*}[h]
\centering
\begin{tabular}{llccc}
\toprule
\multicolumn{5}{c}{\textbf{$\Delta Prob$}} \\
\toprule
\textbf{Model} & \textbf{Method} & \textbf{t-statistic} & \textbf{p-value} & \textbf{Cohen's d} \\
\midrule
Gemma 2 2B & Post-MLP F vs. Post-Att F & 152.29 & $<0.001$ & 1.24 \\
Gemma 2 2B & Post-MLP F vs. MLP F & 56.95 & $<0.001$ & 0.46 \\
Gemma 2 2B & Post-MLP F vs. Neurons & 138.18 & $<0.001$ & 1.13 \\
\midrule
Gemma 2 9B & Post-MLP F vs. Post-Att F & 95.91 & $<0.001$ & 0.78 \\
Gemma 2 9B & Post-MLP F vs. MLP F & 35.27 & $<0.001$ & 0.29 \\
Gemma 2 9B & Post-MLP F vs. Neurons & 146.42 & $<0.001$ & 1.20 \\
\bottomrule
\toprule
\multicolumn{5}{c}{\textbf{$IS$}} \\
\toprule
\textbf{Model} & \textbf{Method} & \textbf{t-statistic} & \textbf{p-value} & \textbf{Cohen's d} \\
\midrule
Gemma 2 2B & Post-MLP F vs. Post-Att F & 28.41 & $<0.001$ & 0.23 \\
Gemma 2 2B & Post-MLP F vs. MLP F & 12.20 & $<0.001$ & 0.10 \\
Gemma 2 2B & Post-MLP F vs. Neurons & 158.37 & $<0.001$ & 1.29 \\
\midrule
Gemma 2 9B & Post-MLP F vs. Post-Att F & 12.55 & $<0.001$ & 0.10 \\
Gemma 2 9B & Post-MLP F vs. MLP F & -12.17 & $<0.001$ & -0.10 \\
Gemma 2 9B & Post-MLP F vs. Neurons & 162.13 & $<0.001$ & 1.32 \\
\bottomrule
\end{tabular}
\caption{Statistical significance test results comparing Post-MLP features with other features or neurons. For each comparison, we report the t-statistic from paired t-tests, corresponding p-value, and Cohen's d effect size.}
\label{appendix:table:statistical_tests:main}
\end{table*}

\section{Paired T-test Results for Main Experiment: Features vs. Neurons}
\subsection{For $\Delta Prob$ and $IS$}
\label{subsection:appenidx:Paired T-test Results for prob and is}
To rigorously validate the comparisons between Post-MLP features and other approaches (Post-Attention features, MLP features, and neurons), we conduct paired t-tests using the full dataset. For both metrics ($\Delta Prob$ and $IS$), we compare Post-MLP features with each alternative method across both models (Gemma 2 2B and Gemma 2 9B). We assess the statistical significance using paired t-tests, as we compare different methods on the same instances.

The results in Table \ref{appendix:table:statistical_tests:main} show varied effect sizes across different comparisons. For $\Delta Prob$, Post-MLP features demonstrate strong advantages over Post-Attention features (Cohen's d: 0.78-1.24) and neurons (Cohen's d > 1.1), while showing more modest advantages over MLP features (Cohen's d: 0.29-0.46). For interpretability ($IS$), we observe particularly strong effects when comparing Post-MLP features with neurons (Cohen's d > 1.2), while comparisons with other feature types show smaller effects ($\lvert\text{Cohen's d}\rvert \leq$ 0.23). All differences are statistically significant (p < 0.001), though the practical significance varies as indicated by the effect sizes.

\subsection{For Monosemanticity}
\label{subsection:appenidx:Paired T-test Results for Monosemanticity}

To rigorously validate the separation phenomenon in activation distributions, we conduct paired t-tests on two types of comparisons: adjacent ratios (e.g., 0\% vs 20\%) and comparisons with the full relation-facts condition (100\%).

The results in Table \ref{appendix: table:monosemanticity_statistics} demonstrate strong and consistent separation patterns, particularly in feature-based representations. For adjacent ratio comparisons, features show large effect sizes (Cohen's d ranging from 0.62 to 5.29) between consecutive ratios, with particularly strong separation in the middle ranges (20\% to 80\%). In contrast, neurons exhibit decreasing effect sizes as the ratio increases, with some comparisons showing small effects (Cohen's d < 0.8) in higher ratios.

When compared against the 100\% baseline, features maintain substantial separation across all ratios (Cohen's d ranging from 1.61 to 13.03), indicating clear distinctions in activation patterns even at high ratios. Neurons, while showing strong separation at lower ratios (Cohen's d > 5.0 for 0\% vs 100\%), demonstrate notably smaller effects at higher ratios (Cohen's d < 1.0 for 80\% vs 100\%). These patterns quantitatively support the superior monosemanticity of features, as they maintain clearer separation between different proportions of relation facts.

\begin{table*}[h]
\centering
\begin{tabular}{llccc}
\toprule
\multicolumn{5}{c}{\textbf{Adjacent Ratio Comparisons}} \\
\toprule
\textbf{Model} & \textbf{Comparison} & \textbf{t-statistic} & \textbf{p-value} & \textbf{Cohen's d} \\
\midrule
Gemma 2 2B Feature & 20 vs. 0 & 27.99 & $<0.001$ & 1.25 \\
Gemma 2 2B Feature & 40 vs. 20 & 118.21 & $<0.001$ & 5.29 \\
Gemma 2 2B Feature & 60 vs. 40 & 57.74 & $<0.001$ & 2.58 \\
Gemma 2 2B Feature & 80 vs. 60 & 67.87 & $<0.001$ & 3.04 \\
Gemma 2 2B Feature & 100 vs. 80 & 35.94 & $<0.001$ & 1.61 \\
\midrule
Gemma 2 2B Neuron & 20 vs. 0 & 50.81 & $<0.001$ & 2.27 \\
Gemma 2 2B Neuron & 40 vs. 20 & 70.10 & $<0.001$ & 3.14 \\
Gemma 2 2B Neuron & 60 vs. 40 & 72.44 & $<0.001$ & 3.24 \\
Gemma 2 2B Neuron & 80 vs. 60 & -2.82 & $0.005$ & -0.13 \\
Gemma 2 2B Neuron & 100 vs. 80 & 12.58 & $<0.001$ & 0.56 \\
\midrule
Gemma 2 9B Feature & 20 vs. 0 & 13.83 & $<0.001$ & 0.62 \\
Gemma 2 9B Feature & 40 vs. 20 & 106.59 & $<0.001$ & 4.77 \\
Gemma 2 9B Feature & 60 vs. 40 & 67.33 & $<0.001$ & 3.01 \\
Gemma 2 9B Feature & 80 vs. 60 & 29.32 & $<0.001$ & 1.31 \\
Gemma 2 9B Feature & 100 vs. 80 & 88.83 & $<0.001$ & 3.97 \\
\midrule
Gemma 2 9B Neuron & 20 vs. 0 & 56.45 & $<0.001$ & 2.53 \\
Gemma 2 9B Neuron & 40 vs. 20 & 29.68 & $<0.001$ & 1.33 \\
Gemma 2 9B Neuron & 60 vs. 40 & 19.10 & $<0.001$ & 0.85 \\
Gemma 2 9B Neuron & 80 vs. 60 & 9.25 & $<0.001$ & 0.41 \\
Gemma 2 9B Neuron & 100 vs. 80 & 17.24 & $<0.001$ & 0.77 \\
\midrule
\bottomrule
\multicolumn{5}{c}{\textbf{Comparisons with 100\% Baseline}} \\
\toprule
\textbf{Model} & \textbf{Comparison} & \textbf{t-statistic} & \textbf{p-value} & \textbf{Cohen's d} \\
\midrule
Gemma 2 2B Feature & 100 vs. 0 & 255.40 & $<0.001$ & 11.43 \\
Gemma 2 2B Feature & 100 vs. 20 & 220.35 & $<0.001$ & 9.86 \\
Gemma 2 2B Feature & 100 vs. 40 & 148.82 & $<0.001$ & 6.66 \\
Gemma 2 2B Feature & 100 vs. 60 & 104.99 & $<0.001$ & 4.70 \\
Gemma 2 2B Feature & 100 vs. 80 & 35.94 & $<0.001$ & 1.61 \\
\midrule
Gemma 2 2B Neuron & 100 vs. 0 & 153.64 & $<0.001$ & 6.87 \\
Gemma 2 2B Neuron & 100 vs. 20 & 112.58 & $<0.001$ & 5.04 \\
Gemma 2 2B Neuron & 100 vs. 40 & 66.82 & $<0.001$ & 2.99 \\
Gemma 2 2B Neuron & 100 vs. 60 & 11.51 & $<0.001$ & 0.51 \\
Gemma 2 2B Neuron & 100 vs. 80 & 12.58 & $<0.001$ & 0.56 \\
\midrule
Gemma 2 9B Feature & 100 vs. 0 & 272.33 & $<0.001$ & 12.18 \\
Gemma 2 9B Feature & 100 vs. 20 & 291.28 & $<0.001$ & 13.03 \\
Gemma 2 9B Feature & 100 vs. 40 & 162.05 & $<0.001$ & 7.25 \\
Gemma 2 9B Feature & 100 vs. 60 & 120.44 & $<0.001$ & 5.39 \\
Gemma 2 9B Feature & 100 vs. 80 & 88.83 & $<0.001$ & 3.97 \\
\midrule
Gemma 2 9B Neuron & 100 vs. 0 & 111.78 & $<0.001$ & 5.00 \\
Gemma 2 9B Neuron & 100 vs. 20 & 74.64 & $<0.001$ & 3.34 \\
Gemma 2 9B Neuron & 100 vs. 40 & 38.81 & $<0.001$ & 1.74 \\
Gemma 2 9B Neuron & 100 vs. 60 & 25.39 & $<0.001$ & 1.14 \\
Gemma 2 9B Neuron & 100 vs. 80 & 17.24 & $<0.001$ & 0.77 \\
\midrule
\bottomrule
\end{tabular}
\caption{Statistical analysis of activation distribution separation. We report t-statistics, p-values, and Cohen's d effect sizes for both adjacent ratio comparisons and comparisons with the 100\% condition. Adjacent ratio comparisons show the separation between consecutive ratios, while baseline comparisons demonstrate the differences from the full relation-facts condition.}
\label{appendix: table:monosemanticity_statistics}
\end{table*}

\section{Synthetic Privacy Dataset Construction and Characteristics}
\label{appendix:privacy-dataset}

Our synthetic privacy dataset comprises 1,500 structured entries of privacy-sensitive information, distributed equally across three relation types: phone numbers (P001), home addresses (P002), and email addresses (P003). Each entry contains a universally unique identifier (UUID), a natural language prompt, the corresponding value, and a relation code. The dataset is specifically designed for privacy-focused machine learning research while ensuring zero risk to individual privacy through complete synthetic generation.

\subsection{Dataset Components}
The dataset construction employs carefully curated component lists to ensure both consistency and variability. We organize our foundational elements into three main categories: identity components, location information, and contact details.

\begin{table*}[t]
\centering
\begin{tabular}{llll}
\toprule
Category & Component Type & Count & Examples \\
\midrule
\multirow{2}{*}{Identity} & First Names & 30 & Alex, Bailey, Casey, Dana, Ellis \\
& Last Names & 30 & Smith, Johnson, Williams, Brown, Jones \\
\midrule
\multirow{3}{*}{Location} & Street Names & 30 & Maple, Oak, Pine, Cedar, Elm \\
& Cities & 30 & Springfield, Rivertown, Lakeside, Hillview \\
& State Codes & 20 & AA, BB, CC, DD, EE \\
\midrule
Contact & Email Domains & 10 & example.com, sample.net, test.org \\
\bottomrule
\end{tabular}
\caption{Dataset generation components.}
\label{tab:components}
\end{table*}

\begin{table*}[t]
\centering
\begin{tabular}{llp{0.3\linewidth}p{0.4\linewidth}}
\toprule
Code & Type & Format Template & Example Query Templates \\
\midrule
P001 & Phone & 555-XXX-XXXX & 
"[name]'s phone number is" \newline
"What is [name]'s phone number?" \newline
"How can I reach [name] by phone?" \\
\midrule
P002 & Address & [Number] [Street] St,\newline [City], [State] [ZIP] & 
"[name]'s home address is" \newline
"Where does [name] live?" \newline
"What is [name]'s residential address?" \\
\midrule
P003 & Email & [firstname].[lastname]\newline[number]@[domain] & 
"[name]'s email address is" \newline
"What's [name]'s email?" \newline
"How can I contact [name] by email?" \\
\bottomrule
\end{tabular}
\caption{Relation types and query templates.}
\label{tab:relation-types}
\end{table*}

\begin{table*}[t]
\centering
\begin{tabular}{llll}
\toprule
Field & Type & Description & Example \\
\midrule
uuid & string & Unique identifier & "550e8400-e29b-41d4-a716-446655440000" \\
sentence & string & Natural language prompt & "Casey Thompson's phone number is" \\
answer & string & Corresponding value & "555-234-5678" \\
relation & string & Relation type code & "P001" \\
\bottomrule
\end{tabular}
\caption{Dataset entry structure.}
\label{tab:entry-structure}
\end{table*}

\subsection{Generation Process}
The dataset generation follows a systematic process to ensure consistency and quality. Names are created by combining first and last names from predefined lists, ensuring unique combinations. Values are generated according to type-specific rules: phone numbers follow the "555-XXX-XXXX" format with random digits, addresses combine random street numbers (1-9999) with component elements, and email addresses merge usernames with random numbers (1-999) and domains.

For each privacy fact, we create six variations of natural language queries, covering both declarative statements and questions. This approach expands our 1,500 unique facts into 9,000 total query-answer pairs. The dataset maintains equal distribution across relation types (500 facts each) and ensures no duplicate entries within each type.

Our quality control process focuses on three key aspects. First, we maintain consistent formatting across all entries to ensure data uniformity. Second, we establish strong referential integrity between names and their associated information to maintain data coherence. Third, we ensure reproducibility through systematic component combination, allowing for dataset regeneration when needed.

\subsection{Research Applications}
This dataset supports various research objectives in privacy-preserving machine learning. It enables thorough model evaluation for information retention and leakage, facilitating the development and evaluation of privacy-protecting mechanisms. The dataset also supports analysis of natural language understanding in the context of structured personal information, while enabling assessment of format learning and consistency in generated content. The synthetic nature of the dataset eliminates privacy concerns while maintaining realistic data patterns and relationships, making it ideal for academic research in privacy-preserving technologies.

\end{document}